\documentclass[10pt,twocolumn,letterpaper]{article}
 
\pdfoutput=1
 
\usepackage{tikz}
\usepackage{iccv}
\usepackage{times}
\usepackage{epsfig}
\usepackage{graphicx}
\usepackage{amsmath}
\usepackage{amssymb}
\usepackage{authblk}

\renewcommand{\vec}[1]{\mathbf{#1}}

\usepackage{color}

\definecolor{darkgreen}{RGB}{0,127,0}
\definecolor{nodecolor}{rgb}{0.7,0.2,0.2}
\definecolor{datacolor}{rgb}{0.2,0.45,0.725}
\definecolor{ilcolor}{rgb}{0.6,0.725,0.35}
\definecolor{darkgreen}{RGB}{0,127,0}

\newcommand{\argmax}{\operatornamewithlimits{arg\,max}}

\usepackage[pagebackref=true,breaklinks=true,letterpaper=true,colorlinks,bookmarks=false]{hyperref}

 \iccvfinalcopy 

\def\iccvPaperID{1129} 

\ificcvfinal\fi
\begin{document}

\title{A Two-Layer Conditional Random Field for the Classification of Partially Occluded Objects}

\author[1]{Sergey Kosov}
\author[2]{Pushmeet Kohli}
\author[1]{Franz Rottensteiner}
\author[1]{Christian Heipke}
\affil[1]{Institute of Photogrammetry and GeoInformation, Leibniz Universitat Hannover, Germany}
\affil[2]{Microsoft Research Cambridge, UK}

\renewcommand\Authands{ and }

\maketitle

\begin{abstract}
Conditional Random Fields (CRF) are among the most popular techniques for image labelling because of their flexibility in modelling dependencies between the labels and the image features. This paper proposes a novel CRF-framework for image labeling problems which is capable to classify partially occluded objects. Our approach is evaluated on aerial near-vertical images as well as on urban street-view images and compared with another methods. 
\end{abstract}

\section{Introduction}

Labeling of image pixels is a classical problem in pattern recognition. Probabilistic models of context such as Markov Random Fields (MRF)~\cite{Li2009} or Conditional Random Fields (CRF)~\cite{Kumar2006} have been increasingly used to model dependencies between labels and/or data at neighbouring image sites. This results in smoothed label images compared to local classifiers. A recent comparison of smooth labelling techniques~\cite{Schindler2012} has shown that smoothing is essential in this context, with CRF performing best among the compared techniqes. 

Labelling techniques usually determine a class label for each pixel of an image. This causes problems if the objects to be detected are partially occluded. For instance, the appearance of streets, sidewalks and buildings may not be clear for a computer if they are largerly occluded by objects such as cars or trees. In remote sensing images, characterized by near-vertical views, this has been known to be a problem for a long time, in particular in the context of automated road extraction. Model-based techniques have tried to overcome this problem by treating such objects as context objects in an ad-hoc manner~\cite{HinzBaumgartner2003, GroteetAl2012}, but a systematic statistical model for dealing with occlusions is still missing. Whereas CRF have been applied successfully to many labelling tasks in computer vision, pattern recognition and remote sensing \cite{Kumar2006, Schindler2012, Schnitzspan2009, ShottonWinn2006}, they also have problems with proper labelling of partially occluded objects, in particular if the occluded objects are those one is actually interested in. In this paper we introduce a two-layered Conditional Random Field ({\em tCRF}), which can handle this problem by explicitly modelling {\em two} class labels for each image site, one for the occluded object and one for the occluding one; in this way, the 3D structure of the scene is explicitly considered in the structure of the CRF. Labelling might also be supported by depth information obtained from image matching. 

Previous work on the recognition of partially occluded objects includes~\cite{Leibe2008}, where the objects in the scene are represented as an assembly of parts. The method is robust to the cases where some parts are occluded and, thus, can predict labels for occluded parts from neighbouring unoccluded sites. However, it can only handle small occlusions, and it does not consider the relations between the occluded and the occlusion objects.
There have been a few attempts to include multiple layers of class labels in CRFs \cite{Kumar2005, Schnitzspan2009, ShottonWinn2006}. However, all these papers also use part-based models where the additional layer does not explicitly refer to occlusions, but encodes another label structure. In \cite{Kumar2005} and \cite{Schnitzspan2009}, multiple layers represent a hierachical object structure, \ie each object on higher level interacts with its smaller parts on lower level. In \cite{ShottonWinn2006}, the part-based model is motivated by the method's potential to incorporate information about the relative alignment of object parts and to model longe-range interactions. However, occluded objects are not explicitly reconstructed. Such a part-based approach is not applicable to objects such as roads in near-vertical views. Roads do not consist of parts having a specific appearance and appearing in a fixed spatial structure. Besides, the spatial structure of such part-based models is not rotation-invariant and, thus, requires the availability of a reference direction (the vertical in images with a horizontal viewing direction), which is not available in remote sensing imagery. As a consequence, methods relying on such a reference direction are not applicable to this class of images. In this respect, the method described in this paper is more general and can be applied to both near-vertical images and images with a horizontal viewing direction. Information about the vertical structure of a scene can be incorporated in scenarios where it makes sense to do so, but it is not a preriquisite for our method to work. 

In~\cite{Yin2007}, MRFs are also expanded by additional layers in the temporal domain, related to the previous and subsequent frames in a video sequence. The interactions between these temporal layers are designed for the detection of moving objects. Occlusions are not dealt within this publication. In~\cite{Gou2012}, occluded areas are recovered by fitting geometrical primitives to large background objects using their visible parts, so the whole classification process is supported with additional frameworks, namely contextual prediction~\cite{Tu2010}, and non-parametric label transfer~\cite{Liu2011}. This work directly adresses the problem of recovering the occluded areas and we show that our method, which consists of only one CRF framework can outperform the reported in \cite{Gou2012} method by accuracy of classification for some classes. It is worth nothing that none of the cited publications use depth information as an additional cue to deal with occlusions.

We solve the problem of labelling partially occluded objects by explicitly considering the 3D structure of the scene. For each image site we have two class labels, one corresponding to an occluded object and the other to the occluding one, using a specific class label to encode that no occlusion occurs. The relations between the two class labels per site and the mutual dependencies between class labels at neighbouring sites in each of the two layers are explicitly modelled. Thus, the information from neighbouring unoccluded objects as well as information from the occluding layer will contribute to an improved labelling of occluded objects. Two layers are sufficient for applications which we focus on, though the principle may be expanded to models with multiple layers. To our knowledge, such a two-layered model has not been applied yet. The interaction model between neighbouring image sites is a contrast-sensitive model which considers the relative frequency of class transitions~\cite{Prasad2006}. The data-dependent terms of our CRF are based on the Random Forest approach~\cite{Breiman2001}. Our method is demonstrated on the task of correctly labelling urban scenes containing crossroads, one of the major problems in road extraction~\cite{RavanbakhshetAl2008}, with the main goal of correctly predicting the class labels of image sites corresponding to the road surface. We also evaluate our method on urban street-view images and compare the results with those achieved in~\cite{Gou2012}, though we will also evaluate the quality of detection for the occluding objects.

\section{Conditional Random Fields (CRF)}
\label{sec:CRF}

We assume an image $\textbf{y}$ to consist of $M$ image sites (pixels or segments) $i\in\mathbb{S}$ with observed data $\textbf{y}_i$, i.e., $\textbf{y} = (\textbf{y}_1, \textbf{y}_2,\dots, \textbf{y}_M)^T$, where $\mathbb{S}$ is the set of all sites. With each site $i$ we associate a discrete class label $x_i$ from a given set of classes $\mathbb{C}$. Collecting the class labels $x_i$ in a vector $\textbf{x} = (x_1, x_2, \dots, x_M)^T$, we can formulate the problem of image classification as finding the label configuration $\hat{\textbf{x}}$ that maximises the posterior probability of the labels given the observations, $p(\textbf{x}|\textbf{y})$, thus $\hat{\textbf{x}} = \argmax_x p(\textbf{x}|\textbf{y})$. 
A CRF is a model of $p(\textbf{x}~|~\textbf{y})$ with an associated graph whose nodes are linked to the image sites and whose edges model interactions between neighbouring sites. Restricting ourselves to CRFs where only pairs of nodes interact, the joint posterior $p(\textbf{x}|\textbf{y})$ can be modelled by~\cite{Kumar2006}:
\begin{equation} 
	p(\textbf{x}~|~\textbf{y}) = \frac{1}{Z}\prod_{i\in\mathcal{S}}\varphi_i(x_i, \textbf{y}) \prod_{i\in\mathbb{S}}\prod_{j\in\mathcal{N}_i}\psi_{ij}(x_{i}, x_{j},\textbf{y}).
	\label{eq:RF_prod_phi}
\end{equation}
In Eq.~\ref{eq:RF_prod_phi}, $\varphi_i(x_i, \textbf{y})$ are the {\em association potentials} linking the observations to the class label at site $i$, $\psi_{ij}(x_{i}, x_{j},\textbf{y})$ are the {\em interaction potentials} modelling the dependencies between the class labels at two neighbouring sites $i$ and $j$ and the data \textbf{y}, $\mathcal{N}_i$ is the set of neighbours of site $i$, and $Z$ is a normalizing constant. Applications of the CRF model differ in the way they define the graph structure, in the observed features, and in the models used for the potentials.

\section{Method}
\label{sec:method}

\subsection{Two-Level Conditional Random Fields}
\begin{figure}[htb]
	 \includegraphics[width=1.00\linewidth]{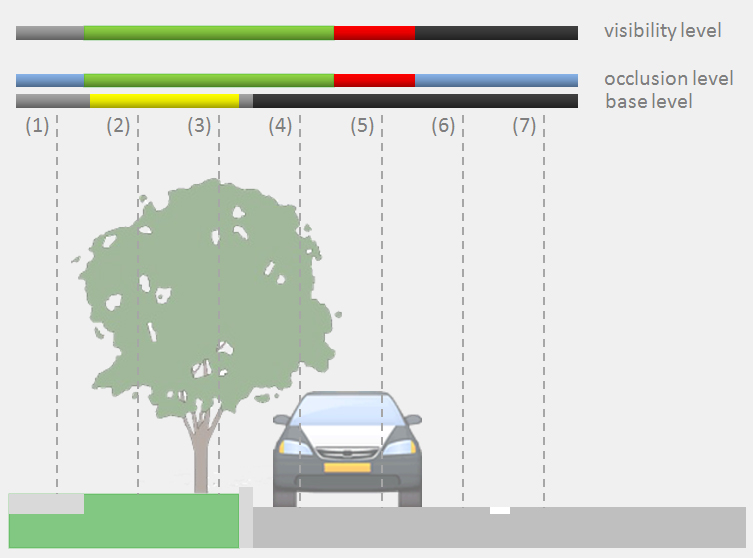} 
   \caption{Two-level set of classes, where labels are represented by colours. Base level: black - street; grey - sidewalk; yellow - grass; Occlusion level: green - tree; red - car; blue - \textit{void}. The visibility level depicts the classes as seen for the sensor (aerial view).}
\label{fig:levels}
\end{figure}

In order to classify partially occluded regions we distinguish objects corresponding to the {\em base level}, \ie the most distant objects that cannot occlude other objects but could be occluded, from objects corresponding to the {\em occlusion level}, \ie all other objects. We separate the objects according to the background-foreground principle: the base level consists of objects such as roads, buildings or grass, whereas the occlusion level includes objects such as cars and pedestrians. 
Consequently, we build a {\em two-level CRF}. Rather than having one label $x_i$ per image site, we determine two such labels $x^{b}_{i}$ and $x^{o}_{i}$, corresponding to the base and occlusion levels, respectively. In general, one occlusion level is sufficient for separating foreground from background. Accordingly, we have two sets of classes, namely $\mathbb{C}^b$ and $\mathbb{C}^o$, corresponding to objects at the base and occlusion levels, respectively, with $x^{b}_{i}\in\mathbb{C}^b$ and $x^{o}_{i}\in\mathbb{C}^o$. Currently, we model $\mathbb{C}^b$ and $\mathbb{C}^o$ to be mutually exclusive, thus $\mathbb{C}^b\bigcap\mathbb{C}^o = \emptyset$. $\mathbb{C}^o$ includes a special class $\textit{void}\in\mathbb{C}^o$ to model situations where the base level is not occluded (Fig.~\ref{fig:levels}). The goal of classification is to determine the most probable values for both $x^{b}_{i}$ and $x^{o}_{i}$ given the data $\textbf{y}$. We model the posterior probability $p(\textbf{x}^b,\textbf{x}^o|\textbf{y})$ directly, expanding the model in Eq.~\ref{eq:RF_prod_phi}:

\begin{eqnarray} 
	\label{eq:CRF}
	p(\textbf{x}^b,\textbf{x}^o|\textbf{y}, \theta) = \frac{1}{Z}\prod_{i\in\mathbb{S}}{\varphi^{b}_{i}(x^{b}_{i},\textbf{y})^{\theta_1}\cdot\varphi^{o}_{i}(x^{o}_{i},\textbf{y})}^{\theta_2}\cdot \nonumber \\ 
	\prod_{i\in\mathbb{S}}\prod_{j\in\mathcal{N}_i}\psi^{b}_{ij}(x^{b}_{i},x^{b}_{j},\textbf{y},\theta_6,\theta_7)^{\theta_3}\cdot\psi^{o}_{ij}(x^{o}_{i},x^{o}_{j},\textbf{y},\theta_6,\theta_7)^{\theta_4}\cdot \\
	\prod_{i\in\mathbb{S}}\xi_i(x^{b}_{i},x^{o}_{i},\textbf{y})^{\theta_5}.\nonumber
\end{eqnarray}

In Eq.~\ref{eq:CRF}, $\theta$ are model parameters: $\theta_1, \theta_2, \ldots, \theta_5\in\theta$ are weights modulating the influence of the individual terms in the classification and $\theta_6, \theta_7\in\theta$ are parameters of the potentials $\psi^{b}_{ij}$ and $\psi^{o}_{ij}$. $\mathcal{N}_i$ is the neighbourhood of  site $i$ (thus, $j$ is a neighbour of $i$). The \textit{association potentials} $\varphi^{b}_{i}$ and $\varphi^{o}_{i}$ link the data $\textbf{y}$ with the class labels $x^{b}_{i}$, $x^{o}_{i}$ of image site $i$. The association potential can be considered as a measure of how likely a site $i$ will take labels $x^{b}_{i}$ or $x^{o}_{i}$ given all image data $\textbf{y}$ and ignoring the effects of other sites in the image.  The \textit{within-level interaction potentials} $\psi^{b}_{ij}$ and $\psi^{o}_{ij}$ model the dependencies between the data $\textbf{y}$ and the labels at two neighbouring sites $i$ and $j$ at the base and occlusion levels, respectively; these potentials correspond to the interaction potentials in Eq.~\ref{eq:RF_prod_phi}. They are related to the probability of how likely the labels of neighbouring sites from one layer $x^{l}_{i}$ and $x^{l}_{j}, l\in\{o,b\}$ are to occur at neighbouring sites given the image data $\textbf{y}$. Finally, in order to link the base and occlusion levels, we define a new \textit{inter-level interaction potential}  $\xi(x^{b}_{i},x^{o}_{i},\textbf{y})$, which models the dependencies between labels from different layers, $x^{b}_{i}$ and $x^{o}_{i}$, and the data $\textbf{y}$. It is a measure of how likely an occlusion of an object at the base level with class label $x^{b}_{i}$ by an object from the occlusion level with class label $x^{o}_{i}$ is to occur, considering the data $\textbf{y}$.  

\begin{figure}[htb]
\center
	\begin{tikzpicture}
		\tikzstyle{data}=[rectangle, draw, text centered, rounded corners, minimum size=2.4em, scale=0.7, datacolor]
		\tikzstyle{var}=[circle, draw, scale=0.7, nodecolor]
		\tikzstyle{darknode}=[fill=nodecolor!20]
		\tikzstyle{darkdata}=[fill=datacolor!20]
		\tikzstyle{edge}=[line width=0.2mm]
		\tikzstyle{note}=[scale=0.75]

		\node[var] (o1) at (0.0,2.8) {$x^{o}_{1}$};
		\node[var] (o2) at (1.2,2.8) {$x^{o}_{2}$};
		\node[var] (o3) at (2.4,2.8) {$x^{o}_{3}$};
		\node[var] (o4) at (3.6,2.8) {$x^{o}_{4}$};
		\node[var] (o5) at (4.8,2.8) {$x^{o}_{5}$};

		\node[var] (b1) at (0.0,1.6) {$x^{b}_{1}$};
		\node[var] (b2) at (1.2,1.6) {$x^{b}_{2}$};
		\node[var,darknode] (b3) at (2.4,1.6) {$x^{b}_{3}$};
		\node[var,darknode] (b4) at (3.6,1.6) {$x^{b}_{4}$};
		\node[var] (b5) at (4.8,1.6) {$x^{b}_{5}$};
		
		\node[data] (d1) at (0.0,0) {$y_1$};
		\node[data] (d2) at (1.2,0) {$y_2$};
		\node[data,darkdata] (d3) at (2.4,0.0) {$y_3$};
		\node[data,darkdata] (d4) at (3.6,0.0) {$y_4$};
		\node[data] (d5) at (4.8,0) {$y_5$};	

		\draw[edge,<->,nodecolor] (b1) edge (b2);
		\draw[edge,<->,nodecolor] (b2) edge (b3);
		\draw[edge,<->,nodecolor] (b3) edge (b4);
		\draw[edge,<->,nodecolor] (b4) edge (b5);

		\draw[edge,<->,nodecolor] (o1) edge (o2);
		\draw[edge,<->,nodecolor] (o2) edge (o3);
		\draw[edge,<->,nodecolor] (o3) edge (o4);
		\draw[edge,<->,nodecolor] (o4) edge (o5);

		\draw[edge,<->,ilcolor] (b1) edge (o1);
		\draw[edge,<->,ilcolor] (b2) edge (o2);
		\draw[edge,<->,ilcolor] (b3) edge (o3);
		\draw[edge,<->,ilcolor] (b4) edge (o4);
		\draw[edge,<->,ilcolor] (b5) edge (o5);

		\draw[edge,->,datacolor] (b1) edge (d1);
		\draw[edge,->,datacolor] (b2) edge (d2);
		\draw[edge,->,datacolor] (b3) edge (d3);
		\draw[edge,->,datacolor] (b4) edge (d4);
		\draw[edge,->,datacolor] (b5) edge (d5);

		\draw[edge,->,datacolor] (o1) to [out=-115,in=180] (d1);
		\draw[edge,->,datacolor] (o2) to [out=-115,in=180] (d2);
		\draw[edge,->,datacolor] (o3) to [out=-115,in=180] (d3);
		\draw[edge,->,datacolor] (o4) to [out=-115,in=180] (d4);
		\draw[edge,->,datacolor] (o5) to [out=-115,in=180] (d5);
		
		\draw[edge,-,dashed,black!40] (-1.4,0.8) to (5.6,0.8);
		
		\node[note,anchor=east,nodecolor] at (-0.8,2.8) {occlusion level};
		\node[note,anchor=east,nodecolor] at (-0.8,1.6) {base level};
		\node[note,anchor=east,datacolor] at (-0.8,0.0) {data};

		\node[note,anchor=west,nodecolor] at (5.4,2.8) {$\vec{x}^o$};
		\node[note,anchor=west,nodecolor] at (5.4,1.6) {$\vec{x}^b$};
		\node[note,anchor=west,datacolor] at (5.4,0.0) {$\vec{y}$};

	\end{tikzpicture}
   \caption{Structure of the tCRF model. The second dimension and additional links between data and labels are omitted for simplicity. Squares and circles correspond to observations and labels, respectively. The dark nodes correspond to a region with occlusion. The graph edges represent dependencies between the nodes.}
\label{fig:crf_struct}
\end{figure}
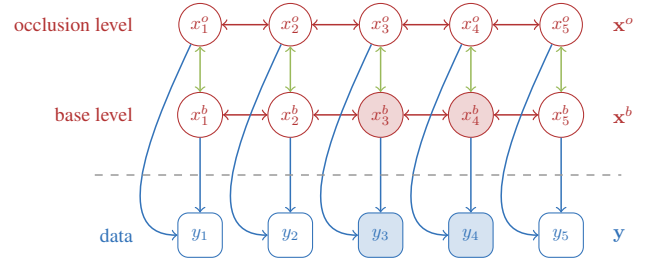

Fig.~\ref{fig:crf_struct} shows the structure of our tCRF model. The dark data nodes represent the input information from  sites with occlusion, \ie where only the occluding object is visible. The reason why we have split the levels is to increase the accuracy of the labelling of occluded regions, \ie to reveal the labels of the dark label nodes in Fig.~\ref{fig:crf_struct}, where the association potentials could not provide the corresponding base level nodes with reliable information because the data corresponding to the base level are not observable.

In a training phase we determine the parameters of the potentials in Eq.~\ref{eq:CRF}, which requires fully labelled training images. The classification of new  images is carried out by maximizing the posterior probability in Eq.~\ref{eq:CRF}. 
The model is very general in terms of the definition of the potentials $\varphi_i$, $\psi_{ij}$ and $\xi_i$. 
Our definitions of the potentials as well as the techniques used for training and inference are described in the subsequent sections. For the sake of simplicity, we will omit the indices $o$ or $b$ in the discussion of the association and the within-layer interaction potentials, assuming the same functional model to be valid for both layers.

%
%
%
%

%
%
\subsection{Association Potential}
\label{sec:pot_func:assoc}
Omitting the superscript indicating the level of the model, the association potentials  $\varphi_{i}(x_{i},\textbf{y})$ are related to the probability of a label $x_{i}$ taking a value $c$ given the  data $\textbf{y}$ by $\varphi_{i}(x_{i},\textbf{y})= p(x_{i}=c|\textbf{f}_i(\textbf{y}))$~\cite{Kumar2006}, where the image data are represented by site-wise feature vectors $\textbf{f}_i(\textbf{y})$ that may depend on all the observations  $\textbf{y}$. 
Note that both the definition of the features and the dimension of the feature vectors $\textbf{f}_i(\textbf{y})$ may vary with the dataset. We use a Random Forest ({\em RF})~\cite{Breiman2001} for the association potentials both of the base and for the occlusion levels, \ie $\varphi^{b}_{i}(x^{b}_{i},\textbf{y})$ and $\varphi^{o}_{i}(x^{o}_{i},\textbf{y})$. 
A RF consists of $N_T$ decision trees that are generated in the training phase. In the classification, each tree casts a vote for the most likely class. If the number of votes cast for a class $c$ is $N_c$, the probability underlying our definition of the association potentials is $p(x_i=c~|~\textbf{f}_i(\textbf{y})) = N_c /N_T$.

\subsection{Within-Level Interaction Potential}
\label{sec:method:WLIP}
The within-level interaction potential  $\psi_{ij}(x_i,x_j,\textbf{y})$ describes how likely the pair of neighbouring sites $i$ and $j$ is to take the labels $(x_i,x_j)=(c,c')$ given the data:  $\psi_{ij}(x_i,x_j,\textbf{y})=p(x_i=c,x_j=c'|\textbf{y})$~\cite{Kumar2006}.
We generate a 2D histogram $h'(x_i,~x_j)$ of the co-occurrence of labels at neighbouring image sites from the training data, \ie $h'(x_i=c,x_j=c')$ is the number of occurrences of the classes $(c,~c')$ at neighbouring sites $i$ and $j$. We scale the rows of $h'(x_i,~x_j)$ so that the largest value in a row will be one  to avoid a bias for classes covering a large area in the training data, which results in a matrix $h(x_i,~x_j)$. 
Our contrast-sensitive definition of $\psi_{ij}(x_i,x_j,\textbf{y})\equiv\psi_{ij}(x_i,x_j,d_{ij})$ is obtained by applying a penalization depending on the Euclidean distance $d_{ij}=\left\|\textbf{f}_i(\textbf{y})-\textbf{f}_j(\textbf{y})\right\|$ of the node feature vectors $\textbf{f}_i$ and $\textbf{f}_j$ to the diagonal elements of $h(x_i,~x_j)$:
\begin{equation}
\psi_{ij}(x_i,x_j,\textbf{y})=\left\{ \begin{array}{rl}
  \theta_6 \cdot e^{-\theta_7 \cdot d_{ij}^2}\cdot h(x_i,~x_j) &\mbox{ if $x_i = x_j$} \\
  h(x_i,~x_j) &\mbox{ otherwise}
       \end{array} \right.
\label{eq:ipotCRF}
\end{equation}
In Eq.~\ref{eq:ipotCRF}, the parameter $\theta_6 \in \theta$ modulates the degree to which the within-level interaction potential favours identical classes at neighbouring sites, whereas $\theta_7 \in \theta$ modulates the contrast-sensitive term. The parameters $\theta_6$ and $\theta_7$ are shared by both inter-level potential functions (base and occlusion levels). As the largest entries of $h_\psi(x_i,~x_j)$ are usually found in the diagonals, a model without the data-dependent term in Eq.~\ref{eq:ipotCRF} would favour identical class labels at neighbouring image sites and, thus, result in a smoothed label image. This will still be the case if the feature vectors $\textbf{f}_i$ and $\textbf{f}_j$ are identical, but large differences between the features will reduce the impact of this smoothness assumption and make a class change between neighbouring image sites more likely. This model differs from the contrast-sensitive Potts model~\cite{BoykovJolly2001} by the use of the normalised histograms $h_\psi(x_i,~x_j)$ in  Eq.~\ref{eq:ipotCRF}. As a consequence, class transitions become more likely, depending on the frequency with which they occur in the training data. Again, the training of the models for the base and the occlusion levels, $\psi^{b}_{ij}(x^{b}_i,x^{b}_j,\textbf{y})$ and $\psi^{o}_{ij}(x^{o}_i,x^{o}_j,\textbf{y})$, respectively, are carried out independently from each other using fully labelled training data. 

\subsection{Inter-Level Interaction Potential}
\label{sec:il-ePot}
The inter-level interaction potential  $\xi_i(x^{b}_{i},x^{o}_{i},\textbf{y})$ describes how likely two variables of site $i$ are to take the labels $(x^{b}_{i}, x^{o}_{i}) = (c, c')$ given the data: $\xi_i(x^{b}_{i},x^{o}_{i},\textbf{y})=p(x^{b}_{i}=c,x^{o}_{i}=c'|\textbf{y})$. Here $c\in\mathbb{C}^b$ and $c'\in\mathbb{C}^o$. We introduce a new set of class labels $\mathbb{C}^i = \mathbb{C}^b \times \mathbb{C}^o$, which encodes all the possible combinations of two labels $c\in\mathbb{C}^b$ and $c'\in\mathbb{C}^o$ by one label $c''\in\mathbb{C}^i$. Thus, the potential function becomes $\xi_i(x^{b}_{i},x^{o}_{i},\textbf{y})=p(\left\{x^{b}_{i}; x^{o}_{i}\right\}=c''~|~\textbf{f}_i(\textbf{y}))$, which is modelled by the RF approach in the same way as described in Sec.~\ref{sec:pot_func:assoc}.

\subsection{Training and Inference}
Exact probabilistic methods for training of a CRF are computationally intractable~\cite{Kumar2006,Vishwanathan2006}. Thus, approximate solutions have to be used. We determine the parameters of the association, within-level interaction and inter-level interaction potentials separately, using only a part of the training data. The association potentials and inter-level interaction potential are trained using the OpenCV implementation of the RF approach~\cite{openCV}. 
For the each class we use the same amount of training samples $N_{samples}$, which are chosen randomly from the training dataset. This results in a total of $N_{samples}\times N_{classes}$ samples that is used for training of both the association and the inter-level interaction potential. The within-level interaction potentials are derived from scaled versions of the 2D histograms of the co-occurrence of class labels at neighbouring image sites in the way described in Sec.~\ref{sec:method:WLIP}, taking into account all image sites in the training data. 
It is a preriquisite of our method that the training data also have two separate layers of labels, one for the base and one for the occlusion layers, respectively. 
The parameters $\theta = \left\{\theta_1, \theta_2, \theta_3, \theta_4, \theta_5, \theta_6, \theta_7 \right\}$ are trained using the Powell search method~\cite{Kramer2010}, an iterative optimisation algortihm that does not require an estimate for the gradient of the objective function. We determine $\theta$ by  maximising the sum $\Omega$ of the diagonal elements of the confusion matrix obtained by classifying the part of the training data that was not used for training the potentials. 
Exact inference is also computationally intractable for CRFs. We use max-product Loopy Belief Propagation, a standard technique for probability propagation in graphs with cycles~\cite{Kolmogorov2006}.

\section{Evaluation}
\label{sec:experiments}

\subsection{Experiment Setup}
As our method requires test data to consist of two separate layers of class labels for the base and occlusion levels, respectively, we only can use datasets providing this information for evaluation. We used the \textit{Vaihingen}\footnote{The Vaihingen data set was provided by the German Society for Photogrammetry, Remote Sensing and Geoinformation (DGPF)~\cite{Cramer2010}.} and the \textit{StreetScene}~\cite{Bileschi2006} datasets for that purpose.
The Vaihingen dataset consists of $1440$ scenes with a size of $250 \times 250$ pixels. Each scene is a colour-infrared (CIR) true orthophoto and a height grid (digital surface model; DSM) generated from wide baseline multiple overlapping airbourne images with a ground sampling distance (GSD) of 8~cm~\cite{KosovISPRS2012}. Both the CIR image and the DSM are geo-coded, and they are defined on the same grid. The reference labels were generated by manually labelling these data in two separate layers. 
The StreetScene dataset consists of $3547$ colour images of $1280 \times 960$ pixels and contains a reference in the form of polygons that also consider hidden object parts and hence could be used to define the two-layered reference required by our method. The Vaihingen data, based on aerial views, are available in a reference frame aligned with the North direction, which is not helpful to structure the scene because roads and buildings (the dominant objects in these data) are not necessarily aligned in North-South or East-West directions. As the original images were taken at the same flying height, all objects appear at a similar scale. On the other hand, for the StreetScenes data, the vertical (y coordinate axis) provides a physically defined reference direction that is clearly related to the scene structure. Furthermore, the distances at which objects are observed vary considerably, so that the scale of objects varies both within and between different scenes.

For the Vaihingen data we chose the nodes of the graphical model to correspond to single pixels, whereas for the StreetScenes data we used image patches of $5\times 5$ pixels. Thus, each graphical model consisted of $250\times 250$ and $256\times 192$ nodes for the two datasets, respectively. The neighbourhood $\mathcal{N}_i$ of an image site $i$ in Eq.~\ref{eq:CRF} (which defines the red edges of the graphical model in Fig.~\ref{fig:crf_struct}) is chosen to consist of the direct neighbours of site $i$ in the data grid. 
The reference of the Vaihingen dataset has six classes: \textit{asphalt} ({\em asp.}), \textit{building} ({\em bld.}), \textit{tree}, \textit{grass}, \textit{agricultural} ({\em agr.}) and \textit{car}, so that $\mathbb{C}^b = \{\textit{asp.}, \textit{bld.}, \textit{grass}, \textit{agr.}\}$ and $\mathbb{C}^o=\{\textit{tree}, \textit{car}, \textit{void}\}$. 
The reference of the StreetScenes dataset has 9 classes: \textit{road}, \textit{sidewalk}, ({\em sdw.}) \textit{bld.}, \textit{store} ({\em str.}), \textit{tree}, \textit{sky}, \textit{car}, \textit{pedestrian} ({\em ped.}) and \textit{bicycle} ({\em bic.}). Since the reference for this dataset is given by polygons, it occures that some image areas are not covered by any polygon. In order to keep our model consistant, we introduce here class \textit{unknown} ({\em unk.}) and mark with it all the uncovered areas at the base level. At the occlusion level, such areas are marked as \textit{void}, So that  $\mathbb{C}^b = \{\textit{road}, \textit{sdw.}, \textit{bld.}, \textit{str.}, \textit{tree}, \textit{sky}, \textit{unk.}\}$ and $\mathbb{C}^o=\{\textit{ped.}, \textit{car}, \textit{bic.}, \textit{void}\}$.

In each test run, 50\% of the images were used for RF training. Our RFs consist of $N_T = 100$ trees of maximal depth $25$. For the training our RF-based potential functions we used $N_{samples}= 10^5$ samples. We used $8.3\%$ of the images for learning the model parameters $\theta$, and the remaining 41.7\% of images for testing. The classification results were compared with the reference; we report the completeness and the correctness (recall and precision) of the results per class as well as the overall classification accuracy~\cite{Rutzinger2009}. 

\subsection{Features}
\label{sec:Features}
The site-wise feature vectors  $\textbf{f}_i(\textbf{y})$ representing the data in the association potentials and inter-level interaction potential depend on the dataset. For the Vaihingen dataset the original data consist of the three colour values of the CIR orthophoto and the associated DSM height for each pixel. The StreetScenes dataset offers only 3-channel colour images. From these original data, we derive the site-wise feature vectors $\textbf{f}_i(\textbf{y})$, each consisting of $N_f$ features. For numerical reasons, all features are scaled linearly into the range $[0; 255]$ and then quantized by 8 bit. 

The features used for both datasets comprise the \textit{image intensity} (\textit{int}), calculated as the average of non-infrared channels and the \textit{saturation} (\textit{sat}) component after transforming the image to the LHS colour space. We also make use of the \textit{variance of intensity} ($\textit{var}_\textit{int}$), the \textit{variance of saturation} ($\textit{var}_\textit{sat}$) and the \textit{variance of gradient} ($\textit{var}_\textit{grad}$) determined from a local neighbourhood of each site $i$ ($7\times 7$ pixels for $\textit{var}_\textit{int}$, $13\times 13$ pixels for $\textit{var}_\textit{sat}$ and $\textit{var}_\textit{grad}$, in both cased evaluated at the original resolution). 

For the Vaihingen dataset includes CIR images, we make use of the \textit{normalized difference vegetation index} (\textit{NDVI}), derived from the near infrared and the red band of the CIR orthophoto \cite{Myneni1995}. For Vaihingen we also determine a digital terrain model (DTM) by applying a morphological opening filter to the DSM with a structural element size corresponding to the size of the largest off-terrain structure in the scene, followed by a median filter with the same kernel size. The DTM is used to derive a {\em normalised DSM} ({\em nDSM})~\cite{WEIDNERFOERSTNER1995}, \ie a model of the height differences between the DSM and DTM. The nDSM describes the relative elevation of objects above ground and its value at image site $i$ is directly used as a feature. Finally, the feature \textit{dist} models the fact that road pixels are usually found in a certain distance either from road edges or road markings. We generate an edge image by thresholding the intensity gradient of the input image. The \textit{dist} feature is the distance of an image site to its nearest edge pixel. The last feature used for the Vaihingen data is the gradient strength of the DSM ($||\nabla DSM||$). 

The StreetScenes dataset has no infra-red channel and no DSM. Nevertheless the classes in  images of this dataset have a strong dependency on the {\em y image coordiante} that reflects the vertical structure of the scenes. For instance, the sky is usually above road and buildings have vertical structure \cite{Yang2011}. Consequently, we use the y coordinate of a node as a feature. This shows that we can incorporate this information when it is helpful (horizontal viewing direction, street scenes), but do not rely on it when it is not available (remote sensing imagery). For similar reasons we make use of histogram of oriented gradients (HOG) features~\cite{DalalTriggs2005} only for the StreetScenes dataset. We calculate the HOG descriptors for cells consisting of $7\times 7$ pixels, using blocks of $2\times 2$ cells for normalization. Each histogram consists of 9 orientation bins ($20^\circ$ per bin). The gradient directions are determined relative to the vertical image axis (which would correspond to a model relative to the North direction in the aerial image case). We extract nine features from the HOG descriptor, namely the  value corresponding to each direction bin ($HOG_0, HOG_1, \dots, HOG_9$). 

For both datasets we make use of multiscale features. That is, the features described above are derived at three different scales. The first scale corresponds to the individual sites, the second and the third are calculated as the average in a local neighbourhoods. For \textit{int}, \textit{sat}, \textit{NDVI} and \textit{nDSM},  these neighbourhoods were chosen to be  $45\times 45$ and $91\times 91$ pixels for the second and the third scales, respectively. For $\textit{var}_\textit{int}$, $\textit{var}_\textit{sat}$, $\textit{var}_\textit{grad}$, \textit{dist}, $||\nabla DSM||$ and the $HOG$ features the neighbourhoods were chosen to be  $10\times 10$ and $100\times 100$ pixels for scales two and three, respectively.

\subsection{Results and Discussion}
\label{sec:Results}
\begin{figure}[htb]
	\center
	\includegraphics[width=1.00\linewidth]{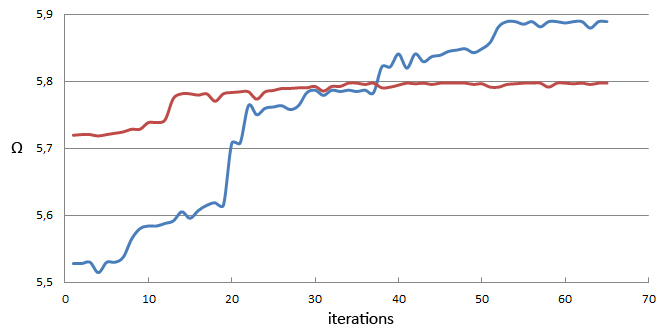} 
	\caption{\label{fig:powell}Convergence of the Powell search method: Red curve: $CRF$; blue curve: $tCRF$.}
\end{figure}

To assess the tCRF model we carried out a number of different experiments. At the first stage we used the \textit{Vaihingen} dataset and performed two experiments: in the first experiment ({\em CRF}), each layer was processed independently, thus the inter-level interaction potentials were not considered. In the second experiment ({\em tCRF}) we use the tCRF model with the inter-level interaction potentials. Fig.~\ref{fig:powell} shows the convergence behaviour of the Powell method for training the parameters $\theta$ in Eq.~\ref{eq:CRF} for both cases. It shows that originally the procedure converges more slowly for the {\em tCRF} method, probably due to a relatively poor initialisation of some parameters, but in the end-iterated state, a larger value of the objective function $\Omega$ can be achieved.

\begin{figure}[tbh]
	\center
	\begin{tabular}{@{}c@{ } @{ }c@{}}
		\includegraphics[width=0.49\linewidth]{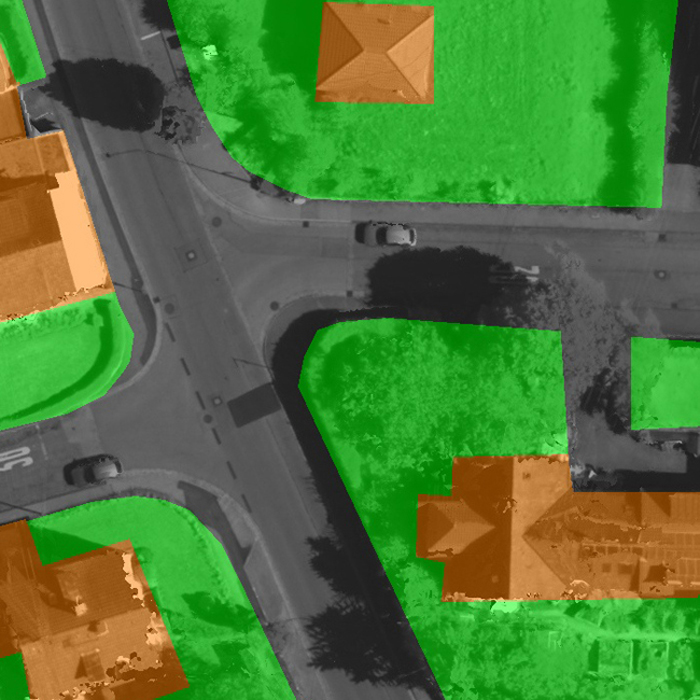} &
		\includegraphics[width=0.49\linewidth]{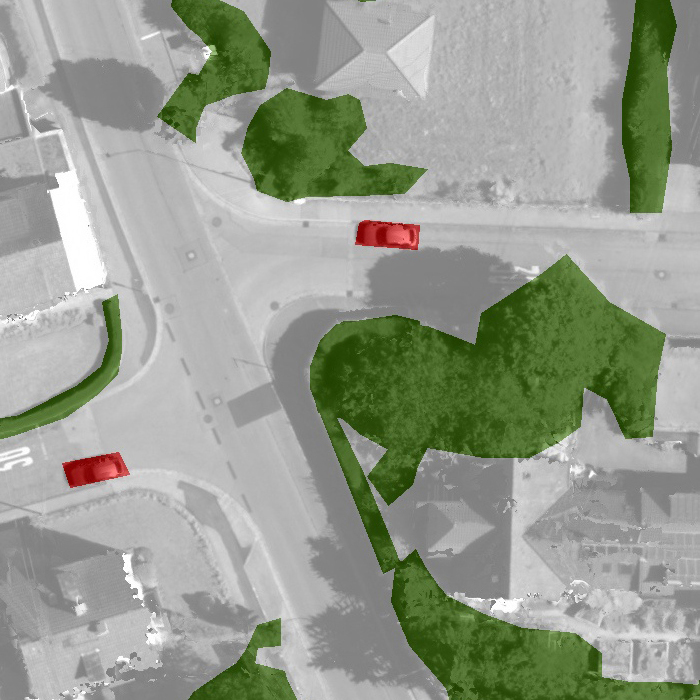} \\
		
		\includegraphics[width=0.49\linewidth]{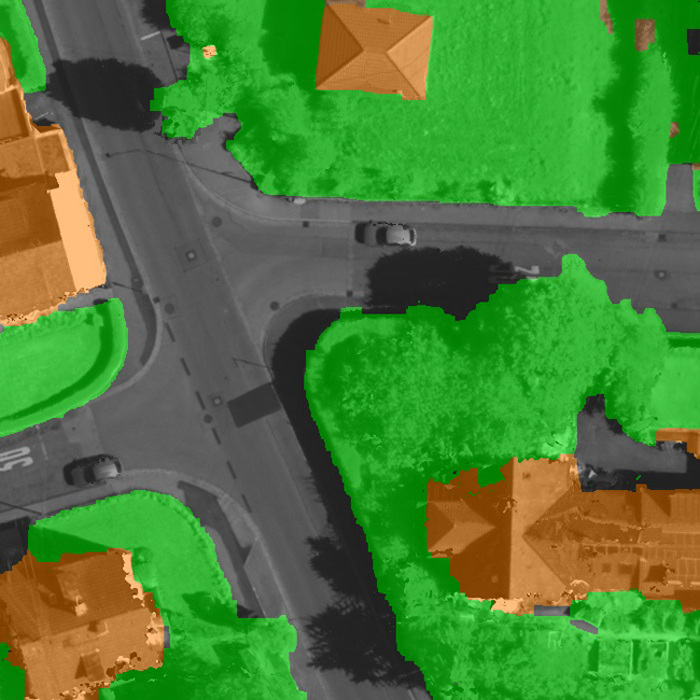} &
		\includegraphics[width=0.49\linewidth]{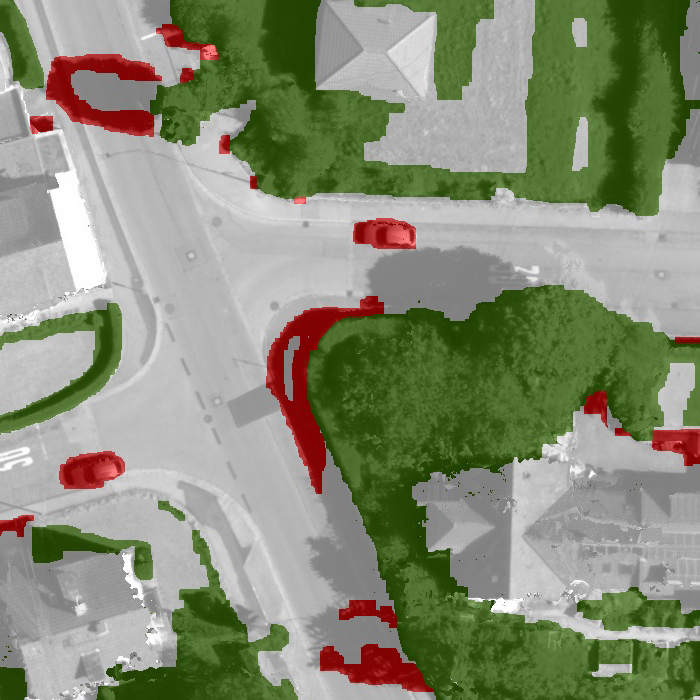} \\
		
		\includegraphics[width=0.49\linewidth]{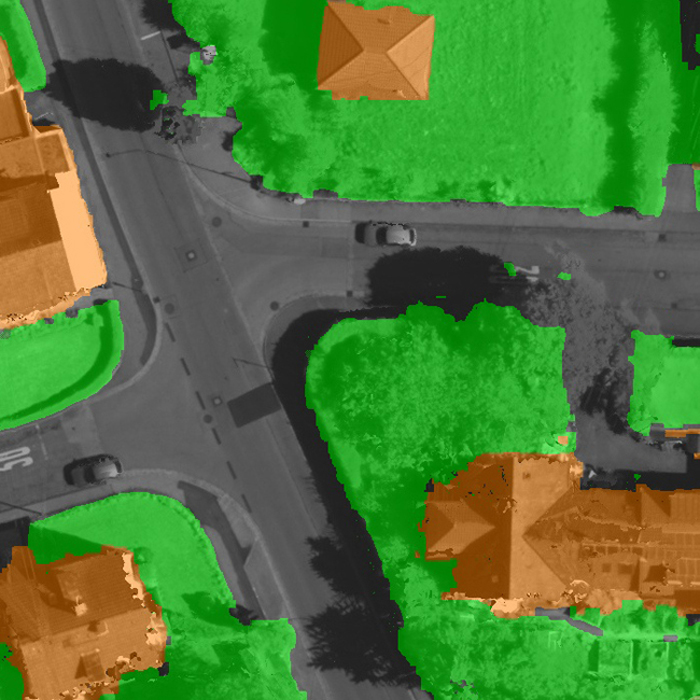} &
		\includegraphics[width=0.49\linewidth]{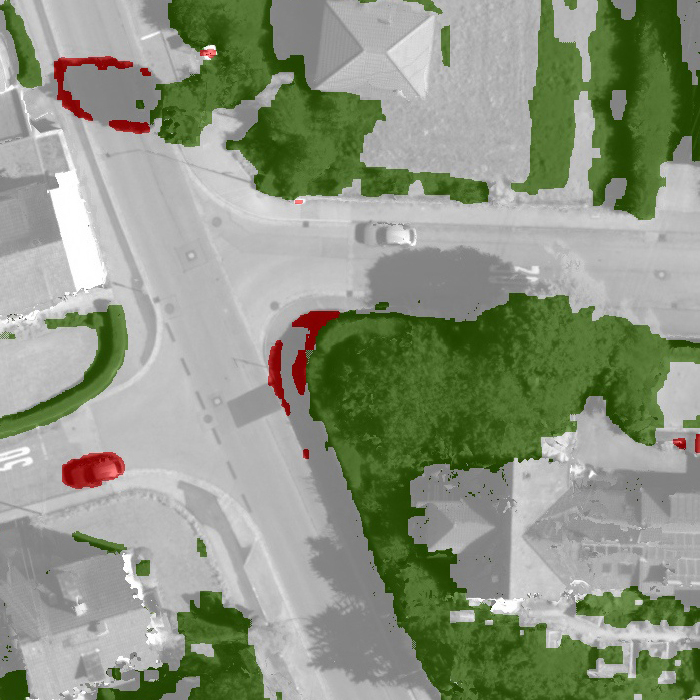} \\
		
		(a) & (b)  \\
	\end{tabular}
	\caption{\label{fig:vaihingen_22}Vaihingen (scene 22). First row: reference, second row: $CRF$, third row: $tCRF$. (a) Base level; (b) Occlusion level. Gray: \textit{asp.}; orange: \textit{bld.}; green: \textit{grass}; beige: \textit{agr.}; white: \textit{void}; darkgreen: \textit{tree}; red: \textit{car}.}
\end{figure}

\begin{figure}[tbh]
	\center
	\begin{tabular}{@{}c@{ } @{ }c@{}}
		\includegraphics[width=0.49\linewidth]{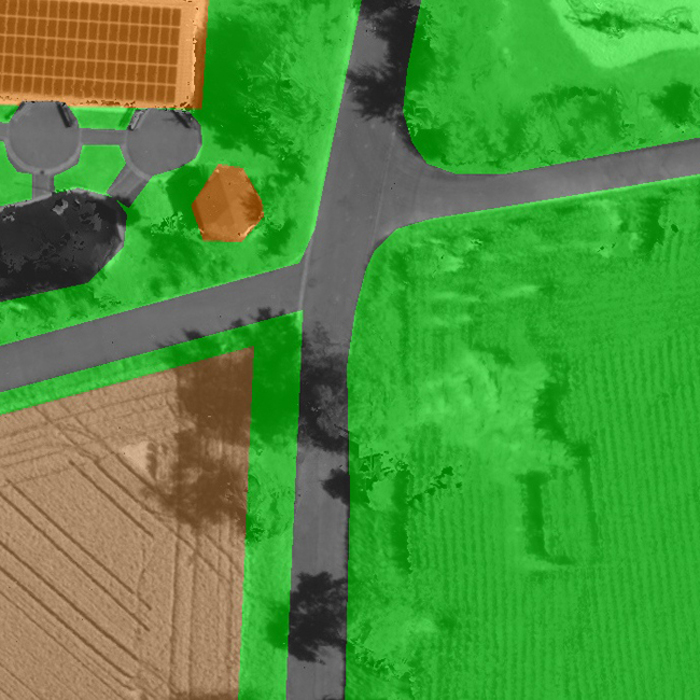} &
		\includegraphics[width=0.49\linewidth]{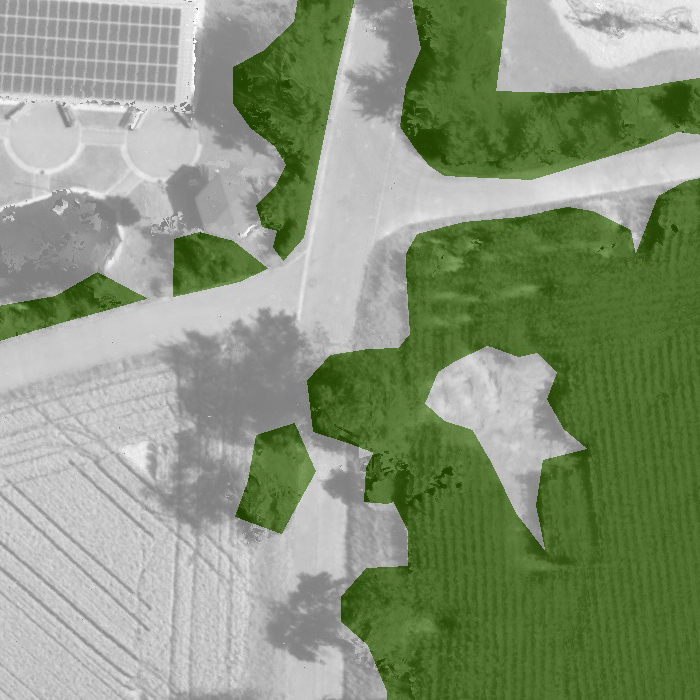} \\
		
		\includegraphics[width=0.49\linewidth]{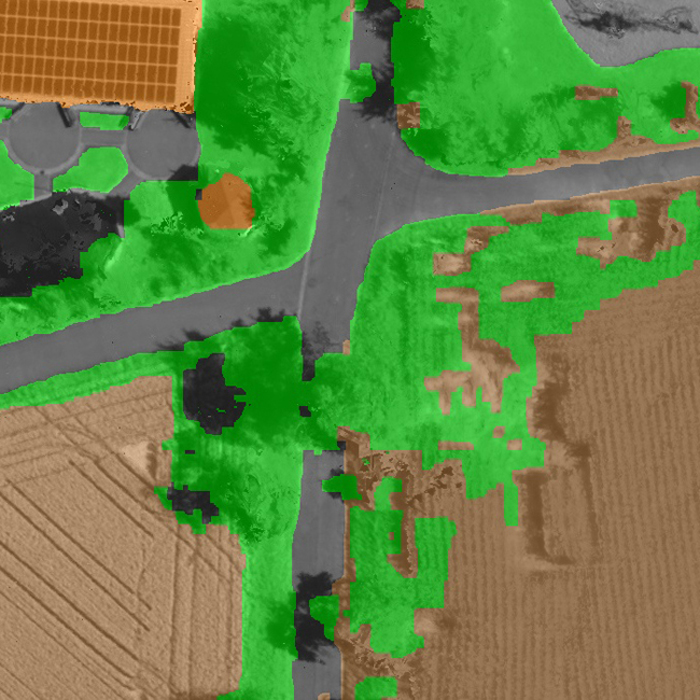} &
		\includegraphics[width=0.49\linewidth]{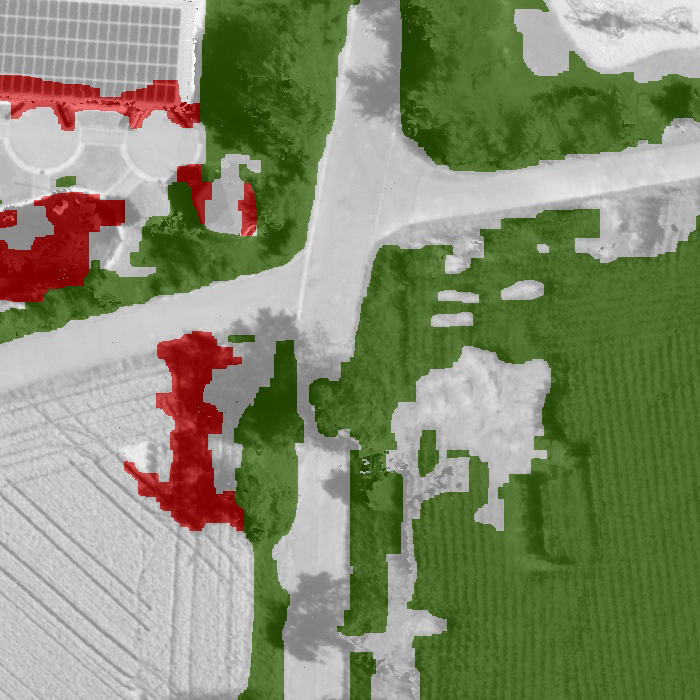} \\
		
		\includegraphics[width=0.49\linewidth]{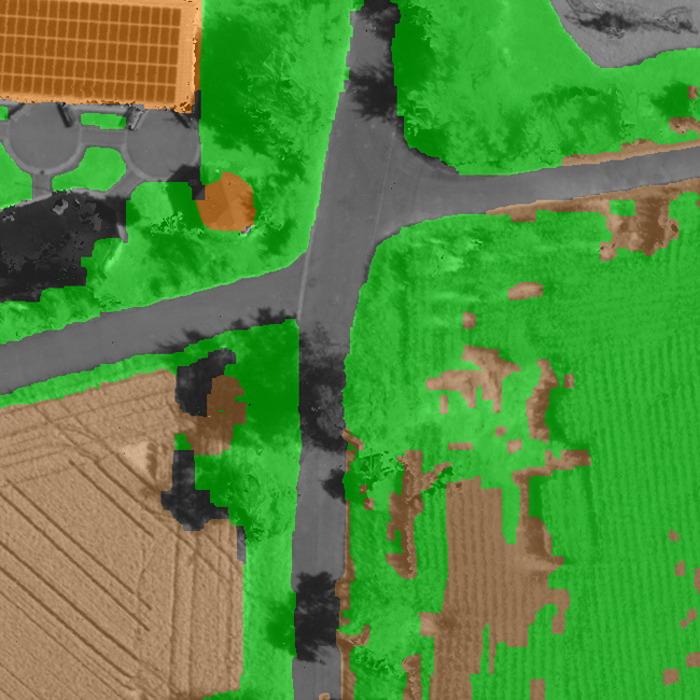} &
		\includegraphics[width=0.49\linewidth]{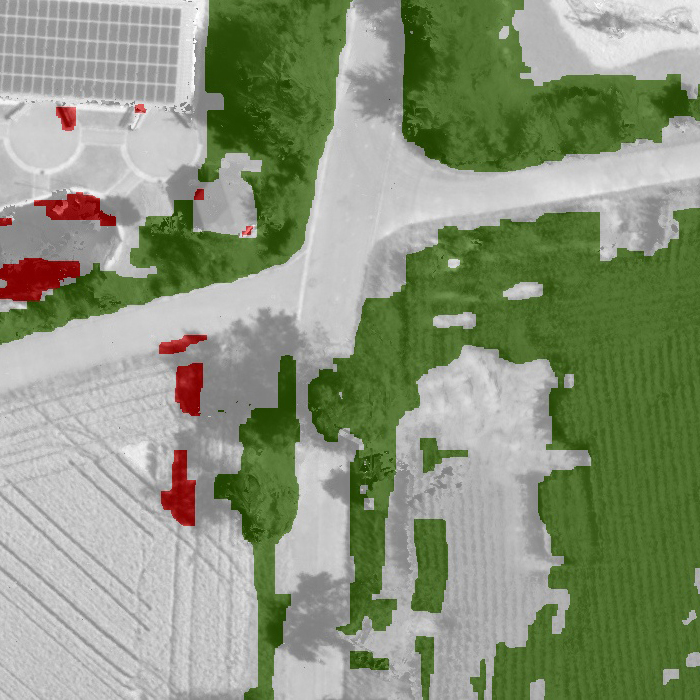} \\
		
		(a) & (b)  \\
	\end{tabular}
	\caption{\label{fig:vaihingen_43}Vaihingen: scene 43. First row: groundtruth, second row: $CRF$, third row: $tCRF$. (a) Base level; (b) Occlusion level. Gray: \textit{asp.}; orange: \textit{bld.}; green: \textit{grass}; beige: \textit{agr.}; white: \textit{void}; darkgreen: \textit{tree}; red: \textit{car}. }
\end{figure}

Fig.~\ref{fig:vaihingen_22} and \ref{fig:vaihingen_43} show the results of the experiments for two Vaihingen scenes. In both figures we can observe that our two-level model considerable improves the road classification in comparison to the state-of-the-art one-layer model. For example, in the right part of the scene in Fig.~\ref{fig:vaihingen_22},  the {\em tCRF} model successfully extracts a road part that is completely occluded with a tree, while {\em CRF} wrongly labels this area as grass. This improvement is possible because the {\em tCRF} models explicitly considers occlusion, the results of the base level receiving information from spatially neighbouring image sites, multi-scale features, and the second layer of labels.
Fig.~\ref{fig:vaihingen_43} also shows how an occluded road can be correctly classified by the {\em tCRF}. In addition, the grass area in the right lower part of the scene is labelled as agricultural by the {\em CRF} model, in spite of the occlusion level saying that this region is covered by trees. Agricultural regions are rarely covered by a forest, and the {\em tCRF} model can use this knowledge (derived from the training data) in order to classify this area correctly. For both scenes we can observe many false positives for the class {\em car}. Their number is reduced considerably by the {\em tCRF} model, though at the cost of a few false negative cars (Fig~\ref{fig:vaihingen_22}). This is also reflected in the quality numbers in Tab.~\ref{tab:evaluation_vaihingen}.

The completeness and the correctness as well as the overall accuracy of the results achieved in these two experiments are shown in Tab.~\ref{tab:evaluation_vaihingen}. Using the {\em CRF} model, the overall accuracy of the classification was 82.6\% for the base level and 80.4\% for the occlusion level. In the second ({\em tCRF}) experiment the overall accuracy for the base level was 86.6\%. The improvement can be attributed by more accurate classification in the occlusion areas (cf.~Fig.~\ref{fig:vaihingen_22} and \ref{fig:vaihingen_43}). From the Tab.~\ref{tab:evaluation_vaihingen}, we can also observe that both the completeness and correctness of car class are still very low. We think that this is due to the fact that cars are relatively small regions and so are described with our features not well enough. The outcome of additional car-detector may correct this situation. Nevertheless our $tCRF$ model has almost double correctness value for cars, then $CRF$ model, while having smaller completeness value. As far as completeness and correctness are concerned, the major improvement is an increased correctness for {\em asp.} and an improved completeness for grass. The class {\em agr.} has a rather low correctness in the model CRF.
For the occlusion level, we observe the best performance when using {\em tCRF}.

\begin{table}[!ht]
\center
	\begin{tabular}{|l||c|c||c|c||}
	\hline
  & \multicolumn{2}{c||}{$CRF$} & \multicolumn{2}{c||}{$tCRF$} \\ 
  & $Cm$. & $Cr$. & $Cm$. & $Cr$.\\  \hline
{\em asp.}  & {\small 80.2 \%} & {\small 90.3 \%} & {\small 85.0 \%} & {\small 87.7 \%} \\ \hline
{\em bld.}  & {\small 86.5 \%} & {\small 78.3 \%} & {\small 85.9 \%} & {\small 82.5 \%} \\ \hline
{\em grass} & {\small 82.7 \%} & {\small 85.5 \%} & {\small 88.3 \%} & {\small 87.8 \%} \\ \hline
{\em agr.}  & {\small 84.1 \%} & {\small 64.4 \%} & {\small 85.4 \%} & {\small 84.2 \%} \\ \hline
{$\textbf{OA}_{base}$} & \multicolumn{2}{c||}{\small \textbf{82.6 \%}} & \multicolumn{2}{c||}{\small \textbf{86.6 \%}} \\  \hline \hline 
{\em \textit{void}}  & {\small 78.1 \%} & {\small 96.9 \%} & {\small 86.8 \%} & {\small 95.7 \%} \\ \hline
{\em tree} & {\small 90.4} & {\small 58.0 \%} & {\small 86.3 \%} & {\small 65.4 \%} \\ \hline
{\em car}  & {\small 72.7} & {\small 11.5 \%} & {\small 47.7 \%} & {\small 19.4 \%} \\ \hline
{$\textbf{OA}_{occl}$} & \multicolumn{2}{c||}{\small \textbf{80.4 \%}} & \multicolumn{2}{c||}{\small \textbf{86.3 \%}} \\  \hline
\end{tabular}
\caption{Completeness (Cm.), Correctness (Cr.) and overall accuracy (OA) of the results for Vaihingen dataset.} 
\label{tab:evaluation_vaihingen}
\end{table}



\begin{figure}[tbh]
	\center
	\begin{tabular}{@{}c@{ } @{ }c@{}}
		\includegraphics[width=0.49\linewidth]{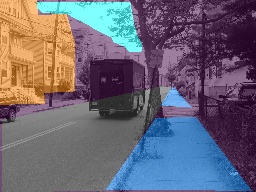} &
		\includegraphics[width=0.49\linewidth]{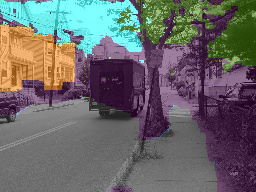} \\
		
		\includegraphics[width=0.49\linewidth]{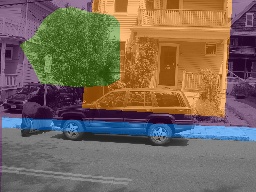} &
		\includegraphics[width=0.49\linewidth]{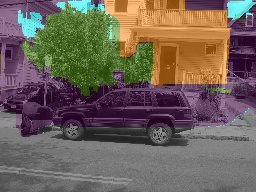} \\
		
		\includegraphics[width=0.49\linewidth]{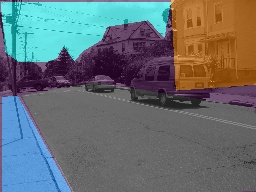} &
		\includegraphics[width=0.49\linewidth]{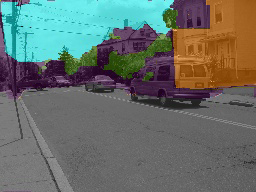} \\
		
		(a) & (b)  \\
	\end{tabular}
		\caption{\label{fig:streetscenes}StreetScenes: examples of base level classification. (a) reference; (b) $tCRF$ classification result. Gray: \textit{road}; blue: \textit{sdw.}; orange: \textit{bld.}; green: \textit{tree}; cyan: \textit{sky}.}
\end{figure}

At the second stage of experiments we used the \textit{StreetScene} dataset and also performed two experiments: {\em CRF} and {\em tCRF} but this time we compare our method with those, reported in~\cite{Gou2012}, namely \textit{Most Confident} ({\em MC}) and \textit{Method of Guo and Hoiem} ({\em GH}). The results are presented in Tab.~\ref{tab:evaluation1} and some classification examples are depicted in Fig.~\ref{fig:streetscenes}.

\begin{table}[!ht]
\center
	\begin{tabular}{|l||c|c|c|c||}
	\hline
  & $CRF$  & $MC$ & $GH$ & $tCRF$ \\ \hline
{\em road} & {\small 90.9 \%} & {\small 92.5 \%} & {\small 93.0 \%} & {\small \textbf{94.5 \%}} \\ \hline
{\em sdw.} & {\small 0.5 \%}  & {\small 28.5 \%} & {\small \textbf{52.5 \%}} & {\small 0.3 \%}  \\ \hline
{\em bld.} & {\small 54.3 \%} & {\small \textbf{90.5 \%}} & {\small 90.0 \%} & {\small 46.6 \%} \\ \hline
{\em str.} & {\small 0.0 \%}  & {\small 0.5 \%}  & {\small \textbf{11.0 \%}} & {\small 0.1 \%}  \\ \hline
{\em tree} & {\small 92.3 \%} & {\small 69.5 \%} & {\small 73.5 \%} & {\small \textbf{92.9 \%}} \\ \hline
{\em sky}  & {\small 77.6 \%} & {\small 68.0 \%} & {\small 79.0 \%} & {\small \textbf{80.4 \%}} \\ \hline
\end{tabular}
\caption{Completeness of the results for \textit{StreetScene} dataset. $CRF$: state-of-the-art 1-layer CRF; $MC$: Most Confident method and $GH$: method of Guo and Hoiem, both reported in~\cite{Gou2012}; $tCRF$ our method.} 
\label{tab:evaluation1}
\end{table}

As we can see from Tab.~\ref{tab:evaluation1} neighter $CRF$ nor $tCRF$ can distinguish \textit{sidewalk} and \textit{store} classes. Newertheless, our $tCRF$ method beats the baseline $GH$ method in terms of classification accuracy for 3 of 6 classes: \textit{road}, \textit{tree}, \textit{sky}.

\section{Conclusion}
\label{sec:Conclusion}

In this paper we have presented a novel approach for considering occlusions in classification based on CRF, the two-level CRF model. Due to its two-level structure it is capable to improve the accuracy of object detection for partially occluded objects. The method was evaluated on the set of airborne- as well as on street-view images and showed a considerable improvement of the overall accuracy in comparison to the classical CRF approach. 
In the future we want to extend our two-level architecture to n-level architecture and apply it to different classes of data. This will include the removal of the restriction that the sets of Classes corresponding to different layer have an empty intersection ($\mathbb{C}^b\bigcap\mathbb{C}^o = \emptyset$). Furthermore, we want to include additional cues to obtain a better classification accuracy for the occlusion level, in particular for the class car.




{\small
\bibliographystyle{ieee}
\bibliography{egbib}

\begin{thebibliography}{10}\itemsep=-1pt

\bibitem{Bileschi2006}
S.~M. Bileschi.
\newblock Streetscenes: Towards scene understanding in still images.
\newblock Technical report, PHD DISSERTATION, MASSACHUSETTES INST. OF
  TECHNOLOGY, 2006.

\bibitem{BoykovJolly2001}
Y.~Boykov and M.~Jolly.
\newblock Interactive graph cuts for optimal boundary and region segmentation
  of objects in n-d images.
\newblock In {\em Proc. ICCV}, volume~I, pages 105--–112, 2001.

\bibitem{Breiman2001}
L.~Breiman.
\newblock Random forests.
\newblock {\em Machine Learning}, 45:5--32, 2001.

\bibitem{Cramer2010}
M.~Cramer.
\newblock The {DGPF} test on digital aerial camera evaluation - overview and
  test design.
\newblock {\em Photogrammetrie-Fernerkundung-Geoinformation}, 2(2010):73--82,
  2010.

\bibitem{DalalTriggs2005}
N.~Dalal and B.~Triggs.
\newblock Histograms of oriented gradients for human detection.
\newblock In {\em Proc. CVPR}, pages 886--893, 2005.

\bibitem{GroteetAl2012}
A.~Grote, C.~Heipke, and F.~Rottensteiner.
\newblock Road network extraction in suburban areas.
\newblock {\em {Photogrammetric Record}}, 27:8–--28, 2012.

\bibitem{Gou2012}
R.~Guo and D.~Hoiem.
\newblock Beyond the line of sight: Labeling the underlying surfaces.
\newblock In {\em ECCV}, pages 761--774, 2012.

\bibitem{HinzBaumgartner2003}
S.~Hinz and A.~Baumgartner.
\newblock Automatic extraction of urban road networks from multi-view aerial
  imagery.
\newblock {\em ISPRS J. Photogramm. \& Rem. Sens.}, 58:83–--98, 2003.

\bibitem{Kolmogorov2006}
V.~Kolmogorov.
\newblock Convergent tree-reweighted message passing for energy minimization.
\newblock {\em IEEE Trans. PAMI}, 28(10):1568--1583, Oct. 2006.

\bibitem{KosovISPRS2012}
S.~Kosov, F.~Rottensteiner, C.~Heipke, J.~Leitloff, and S.~Hinz.
\newblock 3d classification of crossroads from multiple aerial images using
  markov random fields.
\newblock In {\em Proc. 22nd ISPRS Congress}, pages XXXIX--B3:479--484, 2012.

\bibitem{Kramer2010}
O.~Kramer.
\newblock Iterated local search with powell's method: a memetic algorithm for
  continuous global optimization.
\newblock {\em Memetic Computing}, 2(1):69--83, 2010.

\bibitem{Kumar2005}
S.~Kumar and M.~Hebert.
\newblock A hierarchical field framework for unified context-based
  classification.
\newblock In {\em Proc. ICCV}, pages 1284--1291, 2005.

\bibitem{Kumar2006}
S.~Kumar and M.~Hebert.
\newblock {Discriminative Random Fields}.
\newblock {\em Int. J. Comput. Vis.}, 68(2):179--201, 2006.

\bibitem{Leibe2008}
B.~Leibe, A.~Leonardis, and B.~Schiele.
\newblock Robust object detection with interleaved categorization and
  segmentation.
\newblock {\em Int. J. Comput. Vis.}, 77:259--289, 2008.

\bibitem{Li2009}
S.~Z. Li.
\newblock {\em Markov Random Field Modeling in Image Analysis}.
\newblock Springer, 3$^{rd}$ edition, 2009.

\bibitem{Liu2011}
C.~Liu, J.~Yuen, and A.~Torralba.
\newblock Nonparametric scene parsing via label transfer.
\newblock {\em IEEE Trans. Pattern Anal. Mach. Intell.}, 33(12):2368--2382,
  2011.

\bibitem{Myneni1995}
R.~B. Myneni, F.~G. Hall, P.~J. Sellers, and A.~Marshak.
\newblock The interpretation of spectral vegetation indexes.
\newblock {\em IEEE-TGARS}, 33:481--486, 1995.

\bibitem{openCV}
{OpenCV}.
\newblock {Machine Learning}.
\newblock {http://docs.opencv.org/modules/ml/doc/ml.html}, Apr. 2013.

\bibitem{Prasad2006}
M.~Prasad, A.~Zisserman, A.~W. Fitzgibbon, M.~P. Kumar, and P.~H.~S. Torr.
\newblock Learning class-specific edges for object detection and segmentation.
\newblock In {\em Proc. Indian Conf. on Comp. Vision, Graphics and Image
  Proc.}, Dec 2006.

\bibitem{RavanbakhshetAl2008}
M.~Ravanbakhsh, C.~Heipke, and K.~Pakzad.
\newblock Road junction extraction from high resolution aerial imagery.
\newblock {\em {Photogrammetric Record}}, 23:405--423, 2008.

\bibitem{Rutzinger2009}
M.~Rutzinger, F.~Rottensteiner, and N.~Pfeifer.
\newblock A comparison of evaluation techniques for building extraction from
  airborne laser scanning.
\newblock {\em IEEE-JSTARS}, 2(1):11--20, 2009.

\bibitem{Schindler2012}
K.~Schindler.
\newblock An overview and comparison of smooth labeling methods for land-cover
  classification.
\newblock {\em IEEE-TGARS}, 50:4534--4545, 2012.

\bibitem{Schnitzspan2009}
P.~Schnitzspan, M.~Fritz, S.~Roth, and B.~Schiele.
\newblock Discriminative structure learning of hierarchical representations for
  object detection.
\newblock In {\em Proc. CVPR}, pages 2238--2245, 2009.

\bibitem{Tu2010}
Z.~Tu and X.~Bai.
\newblock Auto-context and its application to high-level vision tasks and 3d
  brain image segmentation.
\newblock {\em IEEE Trans. PAMI}, 32(10):1744--1757, Oct. 2010.

\bibitem{Vishwanathan2006}
S.~V.~N. Vishwanathan, N.~N. Schraudolph, M.~W. Schmidt, and K.~P. Murphy.
\newblock Accelerated training of conditional random fields with stochastic
  gradient methods.
\newblock In {\em {Proc. 23$^{rd}$ ICML}}, pages 969--976, 2006.

\bibitem{WEIDNERFOERSTNER1995}
U.~Weidner and W.~F{\"o}rstner.
\newblock Towards automatic building reconstruction from high resolution
  digital elevation models.
\newblock {\em ISPRS J. Photogramm. \& Rem. Sens.}, 50(4):38--49, 1995.

\bibitem{ShottonWinn2006}
J.~Winn and J.~Shotton.
\newblock The layout consistent random field for recognizing and segmenting
  partially occluded objects.
\newblock In {\em Proc. CVPR}, 2006.

\bibitem{Yang2011}
M.~Yang and W.~F{\"o}rstner.
\newblock Regionwise classification of building facade images.
\newblock In {\em Photogrammetric Image Analysis}, volume 6952 of {\em LNCS},
  pages 209--220. Springer, 2011.

\bibitem{Yin2007}
Z.~Yin and R.~T. Collins.
\newblock Belief propagation in a 3d spatio-temporal mrf for moving object
  detection.
\newblock In {\em Proc. CVPR}, 2007.

\end{thebibliography}
}

\end{document}